\documentclass[lettersize,journal]{IEEEtran}
\usepackage{amsmath,amsfonts}
\usepackage{algorithmic}
\usepackage{algorithm}
\usepackage{array}
\usepackage[caption=false,font=normalsize,labelfont=sf,textfont=sf]{subfig}
\usepackage{textcomp}
\usepackage{stfloats}
\usepackage{url}
\usepackage{verbatim}
\usepackage{graphicx}
\usepackage{cite}
\usepackage{enumitem}
\usepackage{svg}
\usepackage{comment}
\usepackage{tabularx}
\usepackage{acronym}
\usepackage{xurl}
\usepackage{cleveref}
\usepackage{siunitx}

\definecolor{darkgreen}{rgb}{0.0, 0.5, 0.0}

\definecolor{lightyellow}{HTML}{FFE699}
\definecolor{red_revision}{HTML}{FF0000}

\acrodef{LLM}{Large Language Model}
\acrodef{AI}{Artificial Intelligence}
\acrodef{ML}{Machine Learning}
\acrodef{CoT}{Chain-of-Thought}
\acrodef{XAI}{Explainable AI}
\acrodef{CQA}{Commonsense Question Answering}
\acrodef{NLP}{Natural Language Processing}
\acrodef{T5}{Text-to-Text Transfer Transformer}
\acrodef{ANOVA}{analysis of variance}

\begin{document}

\title{Honey, I Shrunk the Language Model: \\ Impact of Knowledge Distillation Methods on Performance and Explainability}

\author{Daniel Hendriks, Philipp Spitzer, Niklas Kühl, and Gerhard Satzger 
\thanks{Corresponding author: Daniel Hendriks (e-mail: daniel.hendriks@kit.edu). This work involved human subjects in its research. Approval of all ethical and experimental procedures and protocols was granted by the ethics commission of the Karlsruhe Institute of Technology (KIT)}
\thanks{Daniel Hendriks, Philipp Spitzer, and Gerhard Satzger are with the Institute for Information Systems (WIN), Karlsruhe Institute of Technology, 76133 Karlsruhe, Baden-Württemberg, Germany. Niklas Kühl is with the Information Systems (WI) Institute, University of Bayreuth, 95440 Bayreuth, Bayern, Germany.}
\thanks{This article has supplementary downloadable material available at  \protect\url{https://github.com/DanielHendriks/llm-distillation} and \protect\url{10.21227/4gsd-p846}, provided by the authors.}}

\markboth{Journal of \LaTeX\ Class Files,~Vol.~14, No.~8, August~2021}%
{Shell \MakeLowercase{\textit{et al.}}: A Sample Article Using IEEEtran.cls for IEEE Journals}

\IEEEpubid{0000--0000/00\$00.00~\copyright~2021 IEEE}

\maketitle

\begin{abstract}
\ac{AI} has increasingly influenced modern society, recently in particular through significant advancements in \acp{LLM}. However, high computational and storage demands of LLMs still limit their deployment in resource-constrained environments. Knowledge distillation addresses this challenge by training a small student model from a larger teacher model. Previous research has introduced several distillation methods for both generating training data and for training the student model. Despite their relevance, the effects of state-of-the-art distillation methods on model performance and explainability have not been thoroughly investigated and compared. In this work, we enlarge the set of available methods by applying critique-revision prompting to distillation for data generation and by synthesizing existing methods for training. For these methods, we provide a systematic comparison based on the widely used Commonsense Question-Answering (CQA) dataset. While we measure performance via student model accuracy, we employ a human-grounded study to evaluate explainability. We contribute new distillation methods and their comparison in terms of both performance and explainability. This should further advance the distillation of small language models and, thus, contribute to broader applicability and faster diffusion of LLM technology.
\end{abstract}


\newlength{\nomenclaturewidth}
\settowidth{\nomenclaturewidth}{LLM} 
\settowidth{\nomenclaturewidth}{NLP} 

\section{Introduction}
\label{sec:introduction}

\IEEEPARstart{A}I has become integral to modern society, profoundly influencing daily life, industries, and innovation. The advent of \acp{LLM} has particularly advanced fields such as \ac{NLP}~\cite{brown_language_2020, radford_language_2019, le_scao_how_2021, jin_good_2021, mitra_orca_2023} for tasks such as question answering and reasoning \cite{wei_chain--thought_2022, yao_tree_2023, saxena_evaluating_2024, parmar_logicbench_2024, hou_towards_2023}. One essential factor contributing to LLM's effectiveness is the size of these models with several hundred billion parameters \cite{minaee_large_2024, chowdhery_palm_2022, tang_distilling_2019, jiao_tinybert_2020}. The substantial size of LLMs presents considerable challenges to their application in low-resource environments, such as mobile phones and edge devices \cite{zhu_survey_2023, kim_robustness-reinforced_2024, hsieh_distilling_2023, zhong_panda_2024, ballout_efficient_2024, dai_improve_2024, dai_beyond_2024}. The development of smaller language models is predicated on research in knowledge distillation, which focuses on the transfer of knowledge from a large model (the teacher) to a smaller one (the student). In this process, the teacher model is employed to generate data, and the student model is subsequently fine-tuned.

Knowledge distillation aims to maintain the performance and capabilities of the teacher while improving deployment ease, energy efficiency, and inference speed \cite{hsieh_distilling_2023, wu_one_2021, li_distilling_2023, heikkila_were_2022, strubell_energy_2019}. The importance of distillation becomes visible when looking at the landscape of high-quality open-source models: language models such as Alpaca \cite{wang_how_2023} and Code Llama \cite{roziere_code_2023} have been trained on data from larger language models. Besides the performance of the student model, explainability is vital in knowledge distillation, as it can help verify if the model has learned meaningful concepts and can effectively convey \IEEEpubidadjcol them to humans \cite{doshi-velez_towards_2017,grobrugge_explainability_2024}. Explainability, a language model's ability to reason its outputs towards a human, is crucial if we want humans to understand and trust outputs from language models and, ultimately, adopt them not only in everyday life but also for high-stakes decisions. In this light, past work has emphasized and studied the aspect of explainability in AI systems via human-grounded studies  \cite{spitzer_effect_2024,kim_textual_2018,park_multimodal_2018,morrison_impact_2024}. However, research has so far neglected how knowledge distillation affects the explainability of language models. 

\begin{figure}[!t]
    \centering
    \includegraphics[width=1\linewidth]{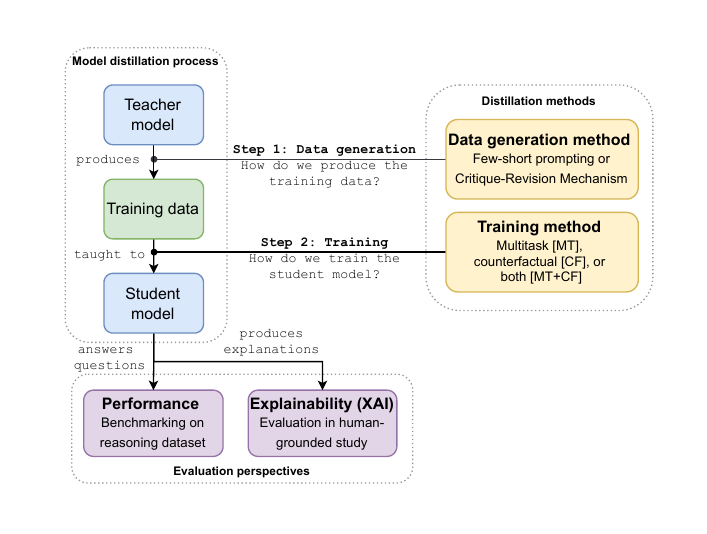}
    \caption{Research model overview: This work introduces novel methods for training data generation and training and establishes approaches to compare performance and explainability of student models.}
    \label{fig:research-model-overview}
\end{figure}

Past work has introduced numerous methods to perform knowledge distillation and improve the final capabilities of the student model. These methods aim to enhance either the training data or the student model. On the one hand, training data is essential, as the performance of student models is significantly affected by the quality of the training data \cite{wang_scott_2023}. To align LLMs' behavior and outputs with human values, such as omitting harmful or biased responses, research has shown that data quality can be enhanced by prompting a language model to improve its own outputs \cite{bai_constitutional_2022}, a strategy we refer to as \emph{critique-revision prompting} \cite{bai_constitutional_2022}. However, the effectiveness of this prompting strategy has not been explored for knowledge distillation. On the other hand, the training of student models is crucial, as demonstrated by recent work by Hsieh et al. \cite{hsieh_distilling_2023}, achieving state-of-the-art performance on various \ac{NLP} benchmarks by training a student via multitask training. Nonetheless, the authors do not explore the impact of multitask training on the student's ability to explain outputs to a human. A second example is the work by Wang et al. \cite{wang_scott_2023} that focuses on counterfactual training resulting in increased faithfulness and consistency in student models, two pillars of a model capable of explaining its outputs. However, their automated \ac{LLM}-based evaluation may lack reliability. Further research using human-centered methods, such as surveys and controlled experiments with participants, is necessary to verify their results. Lastly, comparing the results of existing studies regarding performance and explainability is challenging due to differences in teacher model selection and student model training settings. For instance, Hsieh et al. \cite{hsieh_distilling_2023} use the proprietary PaLM model \cite{chowdhery_palm_2022}, whereas Wang et al. \cite{wang_scott_2023} employ GPT-neox \cite{black_gpt-neox-20b_2022}. Overall, the shortcomings of (1) the lack of evaluation from the view of performance and explainability and (2) the lack of comparability between results motivate our work and give rise to the following two research questions: 

\begin{enumerate}[label=RQ\arabic*:, left=0pt, ]
    \item How does critique-revision prompting affect data quality and consequently the performance and explainability of student models?
    \item How do current training methods affect the performance and explainability of student models?
\end{enumerate}

To address our research questions, we systematically compare distillation methods to evaluate their impact on performance and explainability and combine them to limit potential trade-offs. The high-level logic of this work's research model is presented in \Cref{fig:research-model-overview}: We generate explanations with the teacher through few-shot prompting and aim to improve them through critique-revision prompting. When training the student models, we utilize multitask training \cite{hsieh_distilling_2023}, counterfactual training \cite{wang_scott_2023}, or a combination of both. Performance is empirically evaluated by measuring the accuracy of student models in answering multiple-choice questions from the \ac{CQA} dataset. The explainability of the student model is assessed by generating natural language explanations and evaluating them in a within-subject study with 117 participants. 

In this work, we make three significant contributions: 
\begin{itemize}
    \item We enrich the \textit{spectrum of available distillation methods}. We apply critique-revision prompting to distillation to improve training data generation. We also implement a new training method combining multitask and counterfactual training. We find that applying critique-revision prompting to generate explanations and the combined training method to produce a student model enhances its ability to explain the provided answers to humans.
    \item We develop a \textit{standardized framework for method comparison}--closing a critical gap in current research where heterogeneous teacher models and training configurations have prevented direct comparisons. This also includes a way to assess explainability with human-grounded studies.
    \item We provide a \textit{comprehensive evaluation} of these distillation methods in terms of both performance and explainability. Our comparison shows that models trained with multitask training perform strongly compared to other methods. In terms of explainability, the student model that excels was trained through a combination of multitask and counterfactual training, with explanations enhanced via critique-revision prompting.
\end{itemize}

\section{Related Work}
\label{sec:related-work}

This section reviews existing research on the knowledge distillation of LLMs, focusing on key methods, challenges, and innovations. We begin by discussing knowledge distillation for LLMs, examining its advantages in terms of model size, efficiency, and applicability to resource-constrained environments. We then explore various approaches to data generation, highlighting how explanations and iterative refinement techniques can enhance distillation outcomes. Finally, we review training methods, such as multitask and counterfactual training, which aim to optimize the distillation process for reasoning tasks. Together, these areas form the foundation for improving the capabilities of smaller models while preserving the strengths of their larger counterparts.

\subsection{Knowledge Distillation for LLMs}

Knowledge distillation is a technique introduced by Hinton et al. \cite{hinton_distilling_2015} to transfer knowledge and skills from a large language model to a smaller one. Subsequent research has expanded upon this idea \cite{gu_minillm_2024, tang_distilling_2019, ballout_efficient_2024}, referring to the larger model as the \emph{teacher} and the smaller model as the \emph{student}. In contrast to the teacher, the smaller student model offers four key advantages arising from its reduced size and, thus, lower computational requirements:
\begin{enumerate}
    \item The student can be deployed in resource-constrained environments such as edge devices, smartphones, or limited hardware \cite{hinton_distilling_2015, yu_erdl_2024}. Enabling new applications, this advantage has also been leveraged extensively in computer vision applications \cite{wang_collaborative_2020, feng_triplet_2019, chen_learning_2017}. 
    \item Users can more easily deploy small language models locally, avoiding the high costs associated with proprietary large language models and protecting sensitive data \cite{xu_survey_2024}. 
    \item Their compact size makes smaller models perform inference more quickly and efficiently, leading to greater environmental sustainability \cite{xu_survey_2024}. 
    \item Small models offer lower latency, making them particularly suitable for time-sensitive applications such as entity recognition in industrial settings \cite{izsak_training_2019}. These advantages collectively motivate the need for effective knowledge distillation methods.
\end{enumerate}

To distill an LLM, researchers have proposed various parameter-efficient techniques, including adapters \cite{He_2021}, low-rank structures \cite{hu2021loralowrankadaptationlarge}, and prompt tuning \cite{guo_lopt_2024, lester_power_2021, zhong_panda_2024}. Nevertheless, for distilling reasoning capabilities, the prevalent paradigm specifically involves generating data using the teacher model and subsequently training the student model on this synthetic data \cite{xu_survey_2024}. If successful, this distillation approach enables the student not only to generate coherent text by accurately predicting tokens \cite{mei_efficiency_2024} but also to mimic the teacher’s reasoning processes and correctly estimate uncertainties \cite{mei_efficiency_2024}. Effective distillation of a language model requires two essential ingredients: (a) context-rich and skill-specific training data \cite{xu_survey_2024}, and (b) an appropriate and effective training method \cite{hsieh_distilling_2023, wang_scott_2023}. Both elements are crucial for the student to acquire new capabilities, such as reasoning, by genuinely learning underlying concepts and logic rather than merely memorizing answers or exploiting correlations. To verify that the student model has indeed learned these concepts and logical reasoning, we do not only evaluate the student’s \emph{performance} in this work, but also its capacity to provide explanations for its answers---an ability that is referred to as \emph{explainability} \cite{Holzinger2019CausabilityAE, burkart2021survey}.

\subsection{Data Generation for Knowledge Distillation}
To obtain diverse and rich training data, researchers have proposed techniques prompting the teacher model to explicitly explain its reasoning process step-by-step \cite{dai_improve_2024, xu_baize_2023, wei_chain--thought_2022}. Existing studies demonstrate that incorporating such explanations into distillation leads to capable student models \cite{hsieh_distilling_2023, wang_scott_2023, magister_teaching_2022, ho_large_2023, fu_specializing_2023, dai_beyond_2024, ma_sci-cot_2023} and that enriching or filtering these explanations further improves student performance \cite{mitra_orca_2023, mukherjee_orca_2023, ballout_efficient_2024}.

In this work, we similarly generate explanations from the teacher model, but additionally evaluate a novel technique to expand and refine the data. Specifically, we prompt the teacher model to critique and subsequently revise its own explanations. Inspired by Constitutional AI \cite{bai_constitutional_2022}, we refer to this iterative refinement as \emph{critique-revision prompting}. Initially, Constitutional AI has employed this prompting strategy to produce training data to develop language models that are more helpful and less harmful; however, the authors note that the prompting strategy can guide data generation toward any desired objective. Here, we leverage this approach explicitly to generate high-quality explanations for effective knowledge distillation.

\subsection{Training Methods for Knowledge Distillation}
An effective training method can reliably and effectively support the student in learning concepts and reasoning patterns through a specialized loss function \cite{hsieh_distilling_2023, wang_scott_2023, gu_minillm_2024}. One such method is \emph{multitask training}, which simultaneously teaches the student both explanation and answering tasks. Researchers have proposed using explanations, in some works referred to as ``rationales'', in knowledge distillation by feeding them as additional inputs during training \cite{rajani_explain_2019, wang_self-consistency_2022}. However, this approach requires a teacher model, thus making deployment more complicated and inference neither more energy-efficient nor faster. To address this, Hsieh et al. \cite{hsieh_distilling_2023} frame the problem as a multitask problem and investigate how this training method improves small language models' performance on reasoning datasets. This approach performs better with fewer training examples than conventional training.
Similarly, \emph{counterfactual training} also instructs the student on two concurrent tasks, but differs in the specific tasks involved. Rather than explanation and answering, counterfactual training guides the model to (a) reason and answer correctly when prompted solely with a question, and (b) intentionally answer incorrectly when provided with both the question and an incorrect explanation. This method is motivated by the goal of improving a model’s robustness and sensitivity to misleading information.

\section{Preliminaries}

\begin{figure*}[!t]
    \centering
    \includegraphics[width=0.75\textwidth]{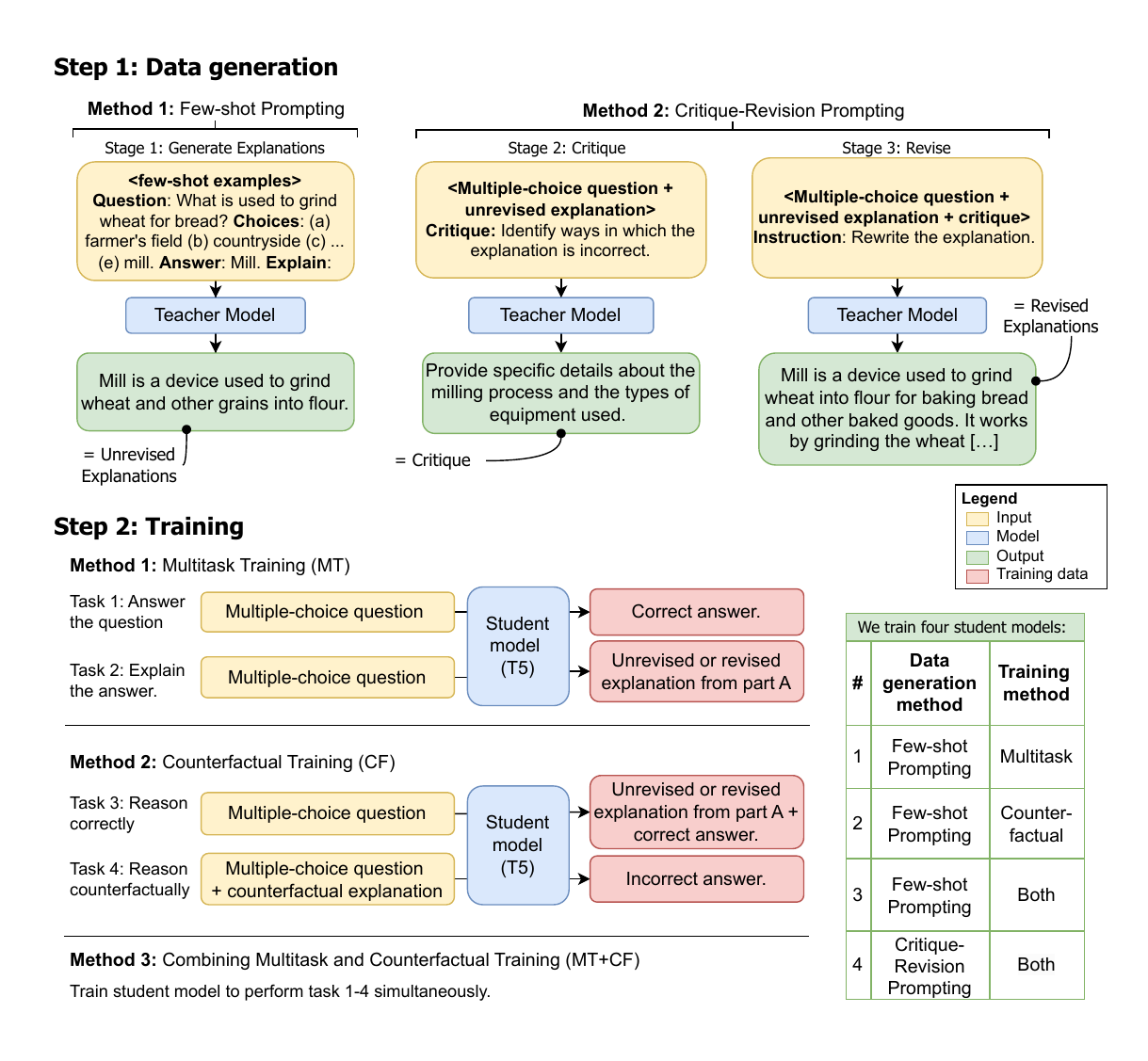}
    \caption{Methods presented in the section on preliminaries are applied in two steps: In \textbf{Step 1}, we generate explanations and then improve them by critiquing and revising the explanation with the teacher.
    In \textbf{step 2}, the explanations are used to fine-tune student models with one of three training methods: multitask training, counterfactual training, or a combination of both. We focus on the four student models shown in the Table for two reasons. On the one hand, they are the most promising candidates in terms of both performance and explainability based on preliminary experiments. On the other hand, to avoid overloading study participants during explainability evaluation, we limit the number of possible student model combinations.}
    \label{fig:research-design}
\end{figure*}

\subsection{Critique-Revision Prompting}
To formalize the described prompting strategy within the context of critique-revision prompting illustrated by \Cref{fig:research-design} on p. \pageref{fig:research-design}, we define the components of the process and specify how each step is handled during the generation and critique phases.

We define the dataset as:

\[
\mathcal{D} = \{(q_i, a_i, e_i, c_i, e_i^\prime)\}_{i=1}^{N}
\]

where \(q_i\) represents the question for the \(i\)-th example, \(a_i\) is the correct answer for the \(i\)-th example, \(e_i\) is the initial explanation generated based on \(q_i\) and \(a_i\), \(c_i\) is the critique of the initial explanation \(e_i\), \(e_i^\prime\) is the revised explanation generated after applying the critique \(c_i\).

In the first of three steps, depicted in \Cref{fig:study-design} on p. \pageref{fig:study-design}, the teacher model is presented with the multiple-choice question \(q_i\), the correct answer \(a_i\), and the instruction to "Explain." The model generates an initial explanation \(e_i\):

\[
e_i = \text{TeacherModel}(q_i, a_i, \text{"Explain"})
\]

In the second step, the teacher model is fed the complete context consisting of \(q_i\), \(a_i\), and \(e_i\), and is instructed to critique \(e_i\). The critique \(c_i\) serves to identify mistakes or areas for improvement in the explanation. This step can be represented as:

\[
c_i = \text{TeacherModel}(q_i, a_i, e_i, \text{"Critique"})
\]

In the third and final step, the teacher model is provided with the complete context \(q_i\), \(a_i\), \(e_i\), and \(c_i\), and is instructed to revise \(e_i\) based on the critique \(c_i\). The revised explanation \(e_i^\prime\) is generated as:

\[
e_i^\prime = \text{TeacherModel}(q_i, a_i, e_i, c_i, \text{"Revise"})
\]

\subsection{Multitask Training}
Multitask training utilizes the \textit{answer} from the CQA dataset and \textit{explanations}, the initial unrevised explanations \(e_i\) or the revised explanations \(e_i^\prime\), to train a student model for the task of answering multiple-choice questions and explaining its predictions. On the one hand, the answers are used as a ground truth to train the student model to answer correctly, by minimizing the following loss function:  

\begin{equation}
\label{equ:standard-loss}
\mathcal{L}_{answer}=\frac{1}{N}\sum_{i=1}^{N}l\left(f\left(q_i\right),a_i\right)
\end{equation}

On the other hand, the student model learns to explain its answers from the explanations according to the following loss function:  

\begin{equation}
\mathcal{L}_{explanation}=\frac{1}{N}\sum_{i=1}^{N}l\left(f\left(q_i\right),e_i\right)
\end{equation}

During training, we insert a label into the prompt (e.g., \textit{[answer]} or \textit{[explain]}) to signal the model to either answering a multiple-choice question or providing an explanation (see~\Cref{fig:research-design}) and update the model's weights based on the gradients from the combined loss:  

\begin{equation}
\mathcal{L}_{multitask} = \mathcal{L}_{answer} + \mathcal{L}_{explanation}
\end{equation}

\subsection{Counterfactual Training}
Performing counterfactual training requires generating counterfactual explanations, i.e., explanations that lead to an incorrect answer, by feeding the teacher model an incorrect answer \({a^\ast}_i\), formalized as follows:

\[
e_i^\ast = \text{TeacherModel}(q_i, a^\ast_i, \text{"Explain"})
\]
Counterfactual training itself follows a similar dual-task structure like multitask training but differs in the objectives assigned to the model. Instead of answering and explaining, counterfactual training consists of two complementary tasks:  (1) answering \textit{and} explaining correctly when given only a question, and (2) answering incorrectly when given a question along with an incorrect explanation.  Formally, we define the loss functions for these two tasks as follows:  

\begin{equation}
\mathcal{L}_{correct}=\frac{1}{N}\sum_{i=1}^{N}l\left(f\left(q_i\right),a_i\right)
\end{equation}

\begin{equation}
\mathcal{L}_{incorrect}=\frac{1}{N}\sum_{i=1}^{N}l\left(f\left(q_i, e^\ast_i\right), a^\ast_i\right)
\end{equation}

The training objective is to minimize the combined loss:  

\begin{equation}
\mathcal{L}_{conterfactual} = \mathcal{L}_{correct} + \mathcal{L}_{incorrect}
\end{equation}

By explicitly training the model to recognize incorrect reasoning and respond accordingly, counterfactual training aims to improve its robustness to misleading or spurious explanations and to enhance model interpretability and generalization \cite{kaushik_learning_2020, wiegreffe_measuring_2021}.  In this work, we hypothesize that combining both multitask and counterfactual leads to improvements in performance and explainability. The idea is to leverage the benefits of both training methods to produce a more capable student model. Based on the previous formalizations, we define the loss of the combined training method as the simple unweighted addition of both losses:
\begin{equation}
\mathcal{L}_{combined} = \mathcal{L}_{multitask} + \mathcal{L}_{counterfactual}
\end{equation}

\section{Methodology}
This section is divided into two main parts. First, we describe how we evaluated the impact of knowledge distillation methods on model performance through an experiment using a standardized dataset and consistent training settings. \Cref{sec:experiment} outlines our datasets, data generation process, training configurations, and evaluation criteria. By systematically analyzing the accuracy of the student models, we quantify the benefits and limitations of different distillation methods.

Second, we show the evaluation of the student models' explainability through a human-based study in which participants evaluated model-generated explanations for multiple-choice questions. \Cref{sec:study-design} details our study design, including participant recruitment and explanation quality criteria. By analyzing human ratings across five key dimensions, we determine how different training and data generation methods affect explanation effectiveness.

\begin{figure*}[t]

    \centering
    \includegraphics[width=0.75\textwidth]{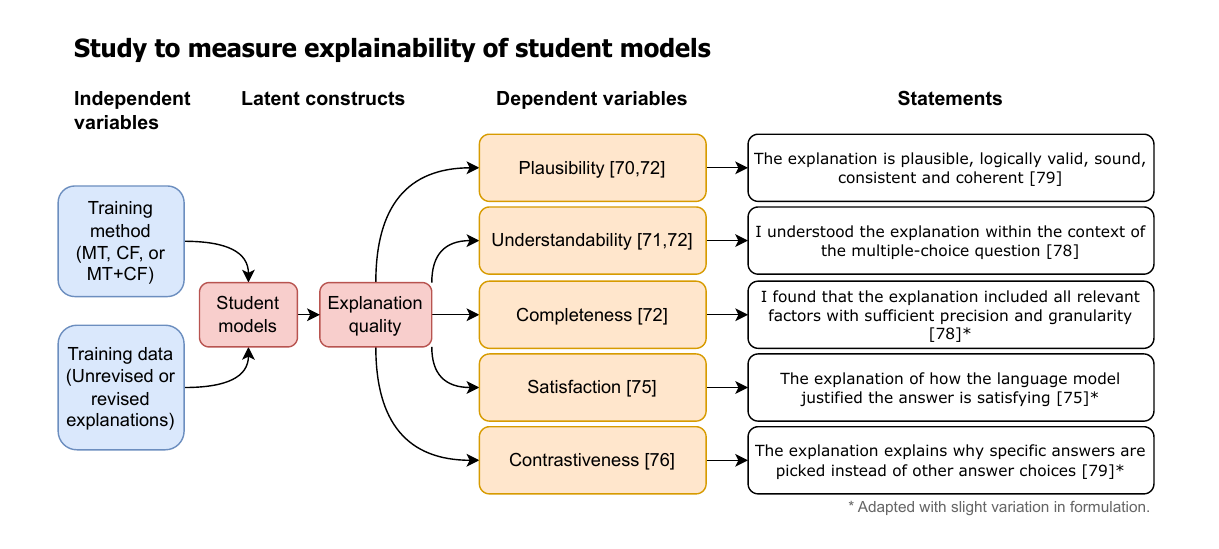}
    \caption{In the within-subject study ($N = 117$), we measure the effect of using four different student models on their explanation quality along five dimensions.}
    \label{fig:study-design}
\end{figure*}

\subsection{Experiment}
\label{sec:experiment}

\subsubsection{Datasets and Data Generation}  
We evaluated the student models on the \ac{CQA} dataset \cite{talmor_commonsenseqa_2019}, consisting of multiple-choice questions with five answer choices, one of which is correct. This dataset, also employed in related studies \cite{hsieh_distilling_2023, wang_scott_2023}, comprises 9,741 training samples and 1,221 test samples. To ensure consistent training and validation across models, we fixed the random seed, enhancing result comparability and reproducibility.  

For explanation generation and critique-revision prompting, we selected \texttt{LLaMA-2-13B} as the teacher model, a state-of-the-art language model at the time of conducting this study \cite{touvron_llama_2023}. Depending on the generation step, we employed either its unaligned or chat-tuned version, following prior research \cite{yang_rlcd_2023, wang_scott_2023}. The settings for critique-revision prompting included 300 new tokens and a temperature of 1. We observe that critiques often exaggerate criticism, and explanations lengthen after applying critique-revision prompting. Initial validation with corrupted samples from the CQA dataset showed that revised explanations add context-relevant information and improve answer differentiation. The generation of counterfactual explanations, which support incorrect answers, is analogous to the initial generation of explanations in critique-revision prompting.

\subsubsection{Training Settings}
We used pre-trained \texttt{T5} architectures as student models: \texttt{T5-base} with 220M million (220M) parameters and \texttt{T5-large} with 770 million (770M) parameters \cite{raffel_exploring_2019}. Trained via the HuggingFace API \cite{wolf_huggingfaces_2019}, student models underwent multitask training, counterfactual training, or a combination of both. Consistent settings were used regardless of the training method: Initial models were trained for up to 10,000 steps; however, since loss and accuracy converged by $5,000$ steps, subsequent models with different random seeds were trained for $5,000$ steps. Training employed the AdamW optimizer with a learning rate of \(5 \times 10^{-5}\) and a maximum generation length of $300$ tokens to prevent truncation. Run on a GRID V100S-16C GPU with 32 GB of memory, reproducibility was ensured by fixing the random seed.

\subsubsection{Evaluation}
We assessed both performance and explainability of student models: For performance, \emph{accuracy} served as the primary metric. Accuracy is defined as the proportion of correctly predicted answers out of the total number of samples: \( \text{accuracy} = \frac{N_{\text{correct}}}{N_{\text{total}}} \), where \( N_{\text{correct}} \) denotes the number of correctly predicted answers, and \( N_{\text{total}} \) represents the total number of evaluated samples. Explainability was assessed through human ratings on explanations generated by student models, following established methodologies \cite{hendricks_generating_2016, park_multimodal_2018, kim_textual_2018}. The data for this evaluation came from a standardized survey where participants rated explanation quality across five dimensions, ensuring a reliable and valid assessment.
\subsection{Human-grounded Study}
\label{sec:study-design}

Our study measured how data generation and training methods affect student models' explainability. We evaluated this by having human participants assess explanations generated by these models for multiple-choice questions from the CQA dataset. After obtaining approval from our university's ethics commission, we recruited 117 participants through Prolific \cite{prolific_prolific_2023}. Each participant assesses 12 explanations (three from each student model) across five quality dimensions, yielding $n = 7,020$ total assessments (117 participants $\times$ 12 explanations $\times$ 5 dimensions). Initially, 120 participants completed the study; three were excluded due to failing attention checks. The participant sample consisted of 59~male, 57~female, and 1~diverse respondents, predominantly from the United Kingdom ($n = 104$). The most represented age groups were 30-34 years ($n = 23$), 25-29 years ($n = 19$), and 40-44 years ($n = 15$). Most participants had higher education ($n = 86$) such as university degrees, vocational university diplomas, A-levels, and other advanced certifications, with 67 employed (including employees, self-employed, and civil servants) and 13 unemployed or seeking employment.

Based on literature on explainability \cite{xie_interpretation_2022, alangari_exploring_2023, zhou_evaluating_2021, chen_rev_2022}, we defined explanation quality through the following five dimensions:

\begin{enumerate}
    \item \textit{Plausibility} is the characteristic of an explanation to be sound and reasonable \cite{oxford_dictionary_oxford_2024}.
    \item \textit{Understandability} is the characteristic of an explanation to be expressed clearly and unambiguously, enabling a human to grasp its meaning easily \cite{oxford_dictionary_oxford_2024}.
    \item \textit{Completeness} is the characteristic of an explanation to describe the reasoning process at an appropriate level of detail and include relevant information \cite{oxford_dictionary_oxford_2024}.
    \item \textit{Satisfaction} is “[…] the degree to which users feel that they sufficiently understand the AI system or process being explained to them.” \cite[p. 3]{hoffman_measures_2023}.
    \item \textit{Contrastiveness} is “[…] the characteristic of an explanation to justify why a prediction was made instead of another.” \cite[p. 18]{carvalho_machine_2019}.
\end{enumerate}

To measure these dimensions, we adapted statements from existing human studies and explanation evaluation guidelines \cite{hase_leakage-adjusted_2020, wang_scott_2023, holzinger_measuring_2020, hebenstreit_collection_2023, mayer_interview_2008, bhattacherjee_social_2012}. Participants rated explanations using a five-point Likert scale after reviewing each multiple-choice question and its corresponding explanation (see~\Cref{fig:study-design}).

To ensure reliability, we used a standardized questionnaire \cite{bhattacherjee_social_2012} with three sections: task introduction, explanation quality evaluations, and socio-demographic data collection. The 117 participants provided 1,404 observations (117 $\times$ 12). We randomized the order of statements and tasks to reduce order effect bias and included only correctly answered explanations to focus on explanation quality rather than answer correctness. We also incorporated attention checks to improve data quality. Participants from Prolific completed our survey in February 2024 and received £2.25 for their participation, contingent on passing the attention check. All participants provided written informed consent prior to participation in this study.

\section{Results and Discussion}
\label{sec:results}

\subsection{Effect on Performance} 
\label{subsec:results-performance}

\begin{figure*}[t]
    \centering
        \subfloat[Performance]{%
        \includegraphics[width=0.75\textwidth]{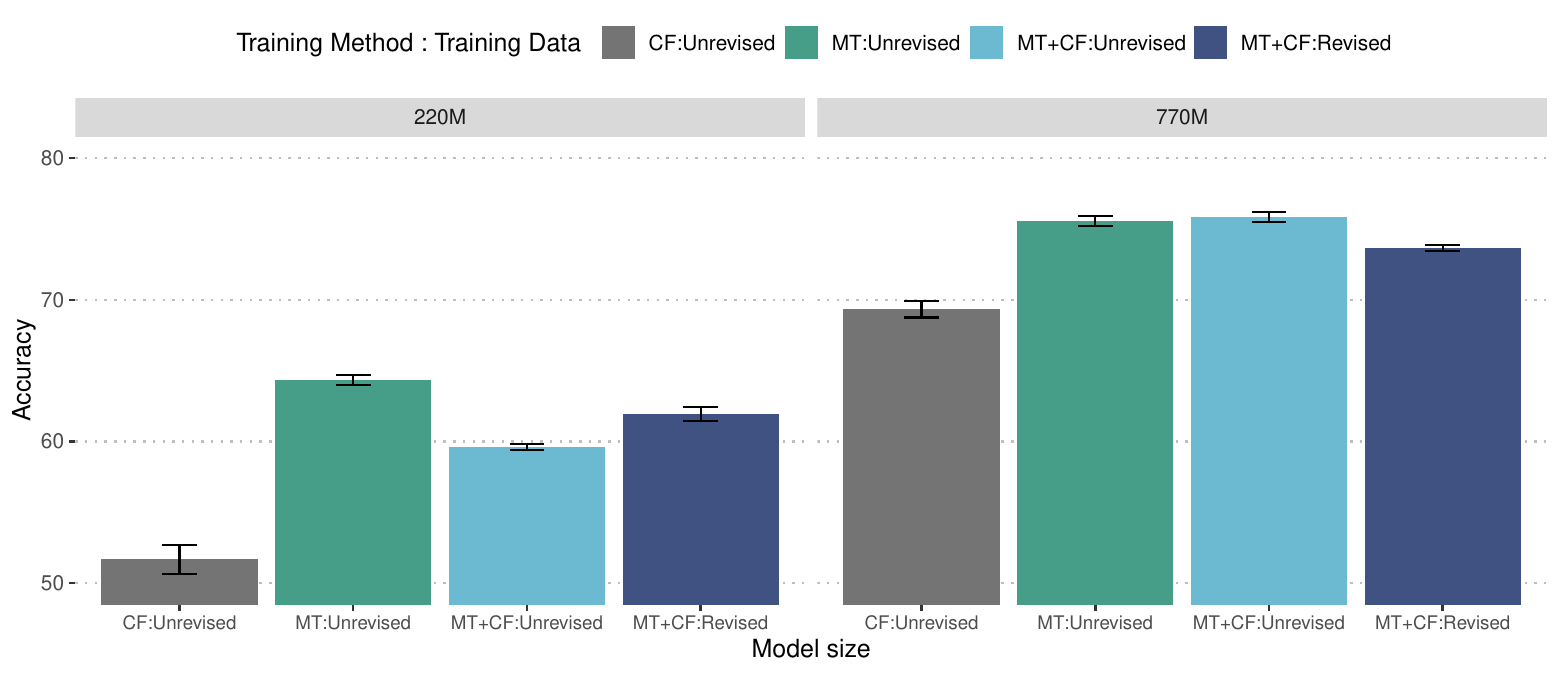}%
    }\vfill
    \subfloat[Explainability]{%
        \includegraphics[width=0.75\textwidth]{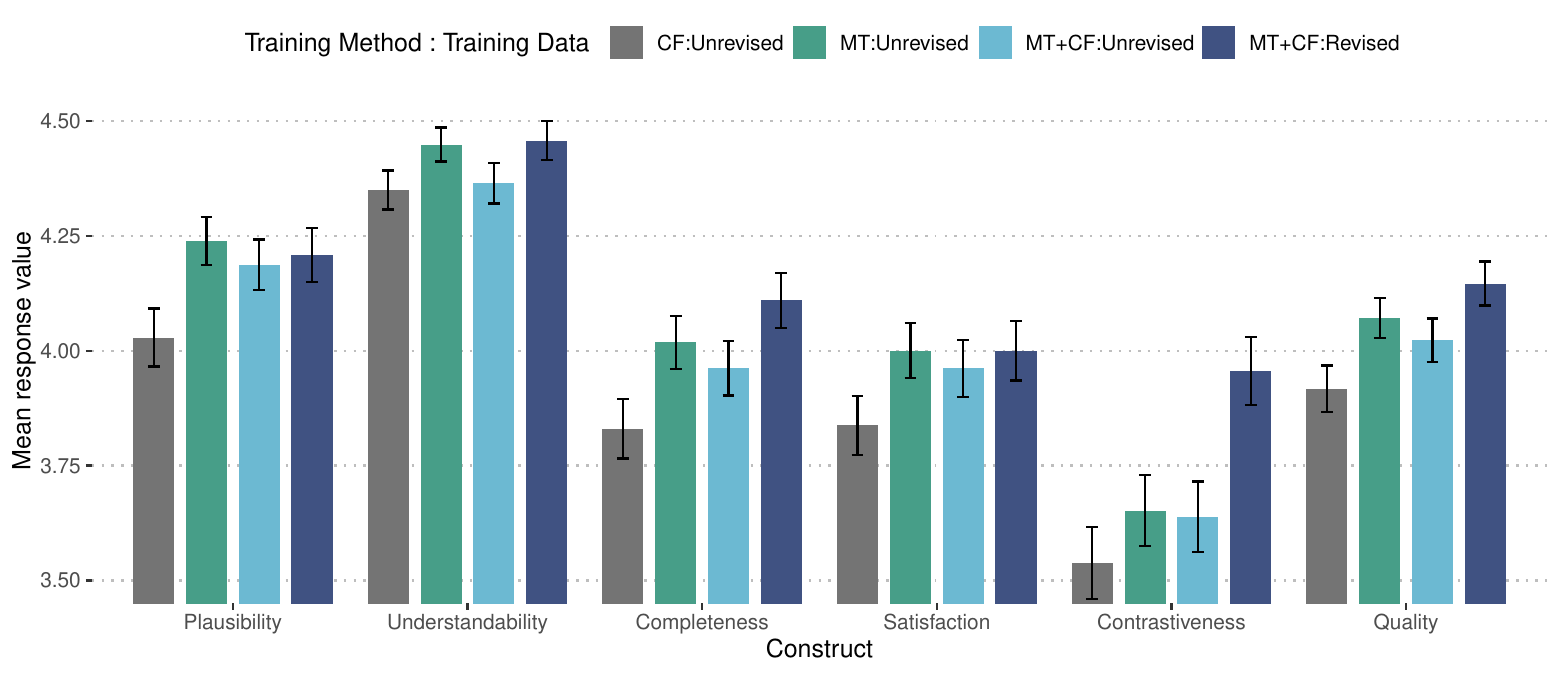}%
    }
    \caption{Comparison of different student models in terms of (a) accuracy performance and (b) explainability measured along the five dimensions of plausibility, understandability, completeness, satisfaction, and contrastiveness. "Quality" as a calculated concept represents the arithmetic average across the five dimensions. Abbreviations: CF for counterfactual training and MT for multitask training.}
    \label{fig:results}
\end{figure*}

The accuracy of models on the \ac{CQA} test dataset is depicted in \Cref{fig:results}(a), with the T5-\text{base} model on the left and the T5-\text{larger} model on the right. Before analysis the performance results from the experiment, outliers identified by the box-plot method, defined as data points lying outside $1.5 \times IQR$, are excluded \cite{paez_exploratory_2022}. For the smaller student models, multitask training with unrevised explanations (MT:Unrevised) achieves the highest mean accuracy, whereas counterfactual training with unrevised explanations yields the lowest performance. For the larger student model, multitask training and the combined training method (MT:Unrevised and MT+CF:Unrevised), both using unrevised explanations, exhibit comparable accuracy, with the combined training method demonstrating a slight advantage. The MT+CF:Revised student models lag behind multitask training on unrevised explanations (MT:Unrevised) in accuracy for both model sizes. In the following paragraphs, the results will be elucidated by conducting a statistical analysis. 

We analyzed the performance assessment results using an \ac{ANOVA}. Before conducting this analysis, we verified the necessary assumptions about data distribution after excluding several outliers. The Shapiro-Wilk test confirmed normality, and the Levene test indicated equal variances, validating the suitability of the \ac{ANOVA} approach. The \ac{ANOVA} results revealed significant performance differences among the student models. Specifically, the type of student model significantly affected performance (\( p = 3.76 \times 10^{-6} \)), demonstrating the impact of critique-revision prompting and training methods. Additionally, model size significantly influenced performance (\( p = 4.96 \times 10^{-14} \)). However, no significant interaction emerged between student model type and model size (\( p = 0.804 \)). This absence of interaction indicates that performance differences due to training and data generation methods are independent of model size, allowing conclusions about these methods to be drawn independently.


\textbf{Critique-revision prompting yields no performance improvement.} Comparing the model MT+CF:Unrevised with MT+CF:Revised allows us to conclude the effect of critique-revision prompting on performance. Based on results from the Tukey-Kramer test presented in \Cref{tab:model_comparisons}, we see that critique-revision prompting negatively affects the larger student model. At the same time, there is no significant effect on the smaller student models' performance. This is unexpected, as critique-revision prompting was intended to enhance data quality and improve model performance. Possibly, the lengthy explanations confused the model, causing performance degradation. Equally surprising is that the effect of critique-revision prompting depends on the student model's size. This dependency might arise because smaller models become more overwhelmed and confused by seeing prolonged explanations during training, resulting in decreased performance or no effect. In contrast, larger models might be able to handle prolonged explanations simply because of their increased number of parameters and, thus, enhanced reasoning capabilities and ability to store information in their parameters.

\textbf{Multitask training performs superior to counterfactual training.} The Tukey-Kramer test shows that the model trained with counterfactual training on unrevised explanations (CF:Unrevised) is consistently outperformed by all other student models for both model sizes, as indicated by the significant difference in accuracy (rows 1 to 3 for the smaller student model and 7 to 9 for the larger student model in \Cref{tab:model_comparisons}). In particular, counterfactual training is also outperformed by models trained with multitask training, leading to the conclusion that multitask training is superior to counterfactual training. The rationale for this observation might be that training a student model to answer multiple-choice questions incorrectly based on counterfactual explanations might confuse the model rather than teaching it more faithful reasoning. In contrast, teaching the student model the correct explanation yields superior performance as seen with multitask training.

\textbf{Combining multitask and counterfactual training yields no performance improvement.} Our findings for the comparison between the student model trained with multitask training on unrevised explanations (MT:Unrevised) and the student model trained with both multitask and counterfactual training on unrevised explanations (MT+CF:Unrevised) are contradictive: for the smaller student models, combining both training methods harms the performance, whereas, for the larger student model, performances are similar (see row 4 and 10). Consequently, overall, across both model sizes, the performance has not improved significantly. Previously, we found that multitask training significantly outperforms counterfactual training. In that light, the lack of performance improvements from combining these methods may result from counterfactual training providing no additional benefit and, therefore, failing to complement multitask training.

\begin{table}[t]
\centering
\caption{Pairwise comparisons of model performance using the Tukey test. A positive estimate indicates that Model 2 outperforms Model 1. Significantly better models are bold.}
\begin{tabular}{lllrS[table-format=2.2]}
\hline
\# & Size & Model 1 & Model 2 & {Estimate} \\ \hline
1  & 220M & CF:Unrevised & \textbf{MT:Unrevised} & 12.66\textsuperscript{***} \\ 
2  & 220M & CF:Unrevised & \textbf{MT+CF:Unrevised} & 7.93\textsuperscript{***} \\ 
3  & 220M & CF:Unrevised & \textbf{MT+CF:Revised} & 10.25\textsuperscript{***} \\ 
4  & 220M & \textbf{MT:Unrevised} & MT+CF:Unrevised & -4.73\textsuperscript{***} \\ 
5  & 220M & MT:Unrevised & MT+CF:Revised & -2.41 \\ 
6  & 220M & MT+CF:Unrevised & MT+CF:Revised & 2.32 \\ \hline
7  & 770M & CF:Unrevised & \textbf{MT:Unrevised} & 6.24\textsuperscript{***} \\ 
8  & 770M & CF:Unrevised & \textbf{MT+CF:Unrevised} & 6.54\textsuperscript{***} \\ 
9  & 770M & CF:Unrevised & \textbf{MT+CF:Revised} & 4.34\textsuperscript{***} \\ 
10 & 770M & MT:Unrevised & MT+CF:Unrevised & 0.30 \\ 
11 & 770M & \textbf{MT:Unrevised} & MT+CF:Revised & -1.90\textsuperscript{*} \\ 
12 & 770M & \textbf{MT+CF:Unrevised} & MT+CF:Revised & -2.20\textsuperscript{**} \\ \hline
\multicolumn{5}{c}{Significance levels: \( ^{***} p<0.001 \), \( ^{**} p<0.01 \), \( ^{*} p<0.05 \), \( ^{\cdot} p<0.1 \)} \\
\hline
\end{tabular}
\label{tab:model_comparisons}
\end{table}

\textbf{Small teacher models are sufficient to produce capable students.} Beyond these findings, we can draw additional conclusions by relating our results to existing literature. Interestingly, we observed that the LlaMA2-13B teacher model produces a student model whose performance is comparable to that obtained from the substantially larger PaLM 540B model. This result challenges the common assumption that increasing teacher model size inherently improves student model performance, highlighting instead the effectiveness of the multitask training approach.

\subsection{Effect on Explainability}
\label{sec:results-human-study} 
\Cref{fig:results}(b) presents the mean values for plausibility, understandability, completeness, satisfaction, contrastiveness, and quality, an average over all previous constructs for each study participant. Overall, \Cref{fig:results} shows that the model MT+CF:Revised produces explanations that are superior in terms of completeness, contrastiveness, and quality. To determine the significance of the differences, we determine an appropriate statistical analysis by assessing the characteristics of the study data: The normality of the six constructs is evaluated via $QQ$-plots indicating an approximately normal distribution; multivariate normality is tested with the Shapiro test (\( p = 9.64 \times 10^{-19} \)) revealing departure from normality; the uni-variate normality Shapiro test demonstrates a significant deviation from a normal distribution for all models and dimensions of explainability; and finally, the Levene test indicates the absence of significant differences in variance among the constructs, thereby suggesting homogeneity. Outliers have been retained in the dataset, as the box-plot method estimates an implausibly high number of outliers to be present in the data, likely due to the ordinal nature of the study data. 

Based on these data characteristics, we perform two analyses: First, to find constructs with significant differences, a series of tests suitable for data of ordinal nature and absence of normality is conducted. Second, two regressions are performed to gain deeper insights into which method impacts student models’ explainability most, with and without demographic data collected during the study. As the construct \textit{quality} is the result of averaging the other constructs, it can be treated as a continuous variable \cite{norman_likert_2010} permitting the use of linear regressions. The procedures and results are described in the subsequent paragraphs.

We use the \ac{ANOVA}-type test to analyze our data, showing that there is a significant difference between the constructs based on the type of student model ($p = 0.003$). Consequently, the rank-based Kruskal-Wallis test is used as a post-hoc test to investigate the differences further. As shown in the \Cref{tab:krusal-wallis-explainability}, significant differences are found for completeness ($p = 0.0081$) and contrastiveness ($p = 0.0004$), justifying the use of Dunn’s test \cite{dinno_nonparametric_2015} to estimate effect sizes. The test shows that the MT+CF:Revised student model significantly outperforms CF:Unrevised, MT:Unrevised, and MT+CF:Unrevised models in terms of contrastiveness, and also surpasses CF:Unrevised significantly in completeness (see~\Cref{fig:results}(b)). Pairwise Vargha and Delaney’s $A$ (VDA) effect size estimates indicate the largest differences between CF:Unrevised and MT+CF:Revised for completeness (VDA $= 0.423$) and contrastiveness (VDA $= 0.404$), although both represent small effect sizes \cite{vargha_critique_2000}. The statistical tests support the initial observation that the MT+CF:Revised student is superior in providing complete and contrastive explanations.

\begin{table}[t]
\centering
\caption{Kruskal-Wallis test results for the five dimensions of explainability. Variables with $p < 0.05$ are bold.}
\begin{tabular}{llS[table-format=1.4]}
\hline
Variable          & n    & {p-value}  \\ \hline
Plausibility      & 1114 & 0.0987   \\ 
Understandability & 1114 & 0.1480   \\ 
\textbf{Completeness}      & 1114 & 0.0081\textsuperscript{**}  \\ 
Satisfaction      & 1114 & 0.1660   \\ 
\textbf{Contrastiveness}   & 1114 & 0.0004\textsuperscript{***}  \\ \hline
\multicolumn{3}{c}{Significance levels: 
  \( ^{***} p<0.001 \), 
  \( ^{**} p<0.01 \), 
  \( ^{*} p<0.05 \), 
  \( ^{\cdot} p<0.1 \)} \\
\hline
\end{tabular}
\label{tab:krusal-wallis-explainability}
\end{table}

\begin{table}[t]
\centering
\caption{Regression results of student models on the formative construct of \emph{quality}. Variables with $p < 0.05$ are bold.}
\begin{tabular}{l S[table-format=1.3] S[table-format=1.3]}
\hline
 & {Estimate} & {Std. Error} \\ \hline
\textbf{Intercept}  & 4.037\textsuperscript{***} & 0.067 \\ 
\textbf{MT:Unrevised} & 0.138\textsuperscript{*} & 0.063 \\ 
MT+CF:Unrevised & 0.109\textsuperscript{$\cdot$} & 0.062 \\ 
\textbf{MT+CF:Revised} & 0.284\textsuperscript{***} & 0.071 \\ 
Explanation length & 0.000 & 0.000 \\ \hline
\multicolumn{3}{c}{Significance levels: 
  \( ^{***} p<0.001 \), 
  \( ^{**} p<0.01 \), 
  \( ^{*} p<0.05 \), 
  \( ^{\cdot} p<0.1 \)} \\
\hline
\end{tabular}
\label{tab:linearregression}
\end{table}

\begin{table}[t]
\centering
\caption{Regression results of student models on the formative construct of \emph{quality}. Only control variables with p-values $< 0.1$ are included for clarity. Baseline categories: ``male'' for gender, ``Civil servant'' for employment status, ``United Kingdom'' for country, ``A-levels/International Baccalaureate/Higher education entrance qualification'' for education, and ``15 to 19 years old'' for age.}
\begin{tabular}{l S[table-format=1.3] S[table-format=1.3]}
\hline
 & {Estimate} & {Std. Error} \\ \hline
\textbf{Intercept}  & 4.496\textsuperscript{***} & 0.470 \\
\textbf{MT:Unrevised} & 0.140\textsuperscript{*} & 0.059 \\
\textbf{MT+CF:Unrevised} & 0.115\textsuperscript{*} & 0.058 \\
\textbf{MT+CF:Revised} & 0.277\textsuperscript{***} & 0.067 \\
\textbf{Gender: Female} & -0.254\textsuperscript{***} & 0.048 \\
Country: No Answer & 0.490\textsuperscript{$\cdot$} & 0.258 \\
\textbf{Country: United States} & -0.541\textsuperscript{***} & 0.090 \\
\textbf{Education: High School Diploma} & -0.209\textsuperscript{*} & 0.085 \\
Education: Junior High Diploma & -0.200\textsuperscript{$\cdot$} & 0.116 \\
\textbf{Education: University Degree} & -0.355\textsuperscript{***} & 0.071 \\
Employment: Pupil/In School & -0.517\textsuperscript{$\cdot$} & 0.312 \\
\textbf{Employment: Unemployed} & 0.412\textsuperscript{***} & 0.124 \\ \hline
\multicolumn{3}{c}{Significance levels: 
  \( ^{***} p<0.001 \), 
  \( ^{**} p<0.01 \), 
  \( ^{*} p<0.05 \), 
  \( ^{\cdot} p<0.1 \)} \\
\hline
\end{tabular}
\label{tab:linearregression_extended}
\end{table}

Before performing the linear regressions, we verify several assumptions about the data. First, we remove 17 outliers identified via the box-plot method, which yields more reliable results for outlier detection now that the data are continuous. Second, we test the homogeneity of variance using the Levene test, which confirms variance homogeneity. Third, we assess normality through $QQ$-plots, which indicate that the data distribution approximates normality. Altogether, these observations justify the use of linear regression. \Cref{tab:linearregression} presents the regression results for student models predicting the formative construct of \emph{quality}. The intercept is significant ($\beta = 4.037$, $p < 0.001$), indicating a baseline quality level for the CF:Unrevised model. Compared to this baseline, the MT:Unrevised condition shows a small but significant positive effect ($\beta = 0.138$, $p < 0.05$), suggesting a modest quality improvement. The MT+CF:Unrevised condition exhibits a marginally significant positive effect ($\beta = 0.109$, $p < 0.1$), indicating a slight potential increase in quality. Notably, the MT+CF:Revised condition demonstrates a significant positive effect ($\beta = 0.284$, $p < 0.001$), clearly highlighting that revisions substantially enhance the quality construct. We control for explanation length, which does not predict explanation quality ($\beta = 0.000$, $p > 0.1$), implying that the length of explanations has no meaningful relationship with perceived quality. These results indicate that counterfactual training alone does not enhance explainability beyond multitask training. 

The regression analysis with control for demographic factors shows that various factors influence the participants' perception of the quality of an explanation (see~\Cref{tab:linearregression_extended}). However, qualitatively, these results yield the same conclusions as the regression without control variables, demonstrating the robustness of our findings.

\textbf{Revising explanations with critique-revision prompting positively impacts the student model’s explainability.} Our analysis indicates that explainability is significantly influenced by revising explanations through critique-revision prompting, which primarily affects contrastiveness and completeness. The difference in effect size between MT+CF:Unrevised and MT+CF:Revised (difference $ = 0.284 - 0.109 = 0.175$) can be attributed to data quality improvements from critique-revision prompting. A possible explanation for this improvement is that revised explanations, being more comprehensive and more differential, directly affect the student model's ability to explain its answer choices (contrastiveness) and provide more comprehensive reasoning (completeness). However, the limited magnitude of these improvements suggests that the student model’s capacity to leverage enhanced explanations may be constrained, potentially by its small size and reasoning capabilities needed to make sense of the complexity introduced through longer, more detailed explanations. In conclusion, while critique-revision prompting did not enhance student model performance, it contributed to explainability. In scenarios where explainability is critical, all three methods, critique-revision prompting, multitask, and counterfactual training, should be combined to produce the student with the best explanation quality.

\textbf{Multitask training improves explainability in contrast to counterfactual training.} Explainability was significantly improved by multitask training compared to counterfactual training, highlighted by the regression results (estimate $ = 0.138$ for MT:Unrevised over the baseline, CF:Unrevised). This finding contradicts the argumentation of Wang et al. \cite{wang_scott_2023}, who posited that counterfactual training enhances explanation factuality. The divergence in findings may stem from the differing evaluation methodologies: this study employed a human-grounded approach, while the previous work relied on functional evaluation via an LLM, highlighting the importance of rigorous assessments of explanation quality from humans.

\textbf{Combining multitask and counterfactual training does not improve the explainability.} In addition, the linear regression analysis shows no significant improvement when using a combination of multitask and counterfactual training without the data quality improvements from critique-revision prompting. Precisely, this can be seen in the regression results in \Cref{tab:linearregression} in the difference between the MT:Unrevised and MT+CF:Unrevised model (difference $ = 0.109 - 0.138 = -0.029$). Evidently, combining multitask and counterfactual training alone does not suffice to achieve this improvement, highlighting that critique-revision prompting primarily drives the observed benefits in explainability.

\section{Conclusion}
\label{sec:conclusion}
In our work, we advance the field of knowledge distillation by (1)~introducing additional methods for data generation and training, (2)~developing a framework for comparing methods on both performance and explainability, and (3)~comprehensively comparing the distillation methods with respect to performance \emph{and} explainability. Our studies yield two main findings: first, regarding performance, multitask training provides a strong student model while maintaining robust levels of explainability. Second, integrating critique-revision prompting with both training methods improves the perceived quality of student model explanations, where the prompting mechanism contributes most to the improvement. These findings strengthen our ability to produce a smaller, more efficient model, promoting sustainability and facilitating deployment in scenarios with limited computational resources while maintaining performance and explainability.

However, this study is not without limitations, which offer the potential for future research. First, by combining multitask and counterfactual training, we did not improve the student models ability to answer and explain multiple-choice questions simultaneously. Thus, future studies could create novel techniques for enhancing training data or methods to boost performance and explainability in student models. Additionally, advanced evaluations of explainability, such as an application-grounded study, might be interesting. Furthermore, this study is limited in the dataset used to evaluate the model -- other researchers should investigate the impact of the investigated methods on performance and explainability in different tasks, use cases, and datasets. Lastly, it would be interesting to investigate the effect of the distillation methods on the student models' ability to generalize to unseen tasks.

\bibliographystyle{IEEEtran}
\bibliography{references}
%


\vfill

\end{document}